# Labelling Vertebrae with 2D Reformations of Multidetector CT Images: An Adversarial Approach for Incorporating Prior Knowledge of Spine Anatomy

Anjany Sekuboyina[1,2] (ME), Markus Rempfler[3] (PhD), Alexander Valentinitsch[2] (PhD), Bjoern H. Menze[1] (PhD), and Jan S. Kirschke[2] (MD)

[1] Department of Informatics, Technical University of Munich, Munich, Germany

[2] Department of Neuroradiology, School of Medicine, Technical University of Munich, Munich, Germany

[3] Friedrich Miescher Institute for Biomedical Engineering, Basel, Switzerland

**Manuscript Type:** Original Research

**Summary Statement:** The proposed fully-convolutional network architecture, Btrfly$_{pe\text{-}eb}$, was trained to learn the a priori knowledge of the spine's shape on two-dimensional maximum intensity projections to locate and identify vertebrae at a rate comparable to prior methods that operated in three dimensions.

**Key Points:**

1. Three-dimensional labelling of vertebrae using two-dimensional sagittal and coronal maximum intensity projections resulted in a computationally lighter, but high performing pipeline, when processed using a 'butterfly' shaped fully convolutional network.
2. Employing spine localization as a pre-processing stage enabled the proposed approach to be applicable to scans of any field-of-view, including complete vertebrae, thus increasing its generalizability to a clinical setting.
3. Enforcing anatomical a priori knowledge (in form of the vertebral arrangement) onto the labelling network using 'adversarial learning' improved the vertebrae identification rate to upwards of 88% on a public benchmarking dataset.

## Abstract

**Purpose:** To use and test a labelling algorithm that operates on two-dimensional (2D) reformations, rather than three-dimensional (3D) data to locate and identify vertebrae.

**Methods:** We improved the Btrfly Net (described by Sekuboyina et al) that works on sagittal and coronal maximum intensity projections (MIP) and augmented it with two additional components: spine-localization and adversarial a priori-learning. Furthermore, we explored two variants of adversarial training schemes that incorporated the anatomical a priori knowledge into the Btrfly Net. We investigated the superiority of the proposed approach for labelling vertebrae on three datasets: a public benchmarking dataset of 302 CT scans and two in-house datasets with a total of 238 CT scans. We employed Wilcoxon signed-rank test to compute the statistical significance of the improvement in performance observed due to various architectural components in our approach.

**Results:** On the public dataset, our approach using the described $\text{Btrfly}_{\text{pe-eb}}$ network performed on par with current state-of-the-art methods achieving a statistically significant ($p < .001$) vertebrae identification rate of 88.5±0.2 % and localization distances of less than 7-mm. On the in-house datasets that had a higher inter-scan data variability, we obtained an identification rate of 85.1±1.2%.

**Conclusion:** An identification performance comparable to existing 3D approaches was achieved when labelling vertebrae on 2D MIPs. The performance was further improved using the proposed adversarial training regime that effectively enforced local spine a priori knowledge during training. Lastly, spine-localization increased the generalizability of our approach by homogenizing the content in the MIPs.

## 1. Introduction

Spine CT is a commonly performed imaging procedure. In this work, we focused on labelling the vertebrae, which is the joint task of locating and identifying the cervical, thoracic, lumbar, and sacral vertebrae in a regular CT spine scan. Labelling the vertebrae has immediate diagnostic consequences (e.g. vertebral landmarks help identify scoliosis, pathological lordosis, and kyphosis). From a modelling perspective, labelling simplifies the downstream tasks of intervertebral disc segmentation or vertebral segmentation.

Previous approaches aimed at labelling broadly fell into two categories: the traditional model-based approaches and relatively recent learning-based approaches. Model-based approaches like Schmidt et al. [1], Klinder et al. [2], and Ma et al. [3] utilized a priori information on the spine structure such as statistical shape models or atlases. Due to their extensive reliance on a priori information, their generalizability was limited. From a machine learning perspective, there have been approaches ranging from regression forests working on context features in Glocker et al. [4], Glocker et al. [5], Suzani et al. [6], a combination of convolutional neural networks (CNN) and random forests in Chen et al. [7], or 3D fully convolutional networks in Yang et al. [8] followed by recurrent neural networks in Yang et al. [9] and Liao et al. [10]. A majority of these approaches work on full 3D multidetector CT scans. Recently, an approach achieved a higher labelling performance by working on 2D maximum intensity projections (MIP) using an architecture referred to as the 'Btrfly' Network proposed by Sekuboyina et al. [11].

In this work, we improved on the Btrfly architecture and extended it with a spine localization module thus making the combination more generalizable. Concurrently, inspired from the generative adversarial learning domain, we investigated an a priori-learning module that enforced spine's anatomical a priori knowledge (or prior) onto the Btrfly network. There have been earlier approaches aimed at prior-learning such as Ravishankar et al. [12] and Oktay et al. [13]. Typically, they consist of two networks, one primary network solving a task such as segmentation and a secondary pre-trained network that learns the shape-of interest and either 'corrects' the primary network's prediction [12] or 'enforces' it on the primary network [13]. Our network was similar to these approaches to the point that it was composed of two components, a labelling network and an adversary as the secondary. However, Btrfly Net was fundamentally different in the training and inference processes compared to the previous works [12,13]. First, the adversary required no pre-training as it was trained along with the labelling network by penalizing it if the labels deviated from the ground truth distribution. Second, since the adversarial loss was used to directly update the weight of the Btrfly net, the adversary was no longer needed during inference.

Thus, our hypothesis was that by incorporating a spine localization stage before vertebral labelling and then adversarially learning an anatomical prior and enforcing it onto the Btrfly network, the labelling performance would be improved while also making the setup generalizable to clinical routines. We describe the use of the prior-encoded Btrfly variant for improved vertebrae identification and labeling compared to the original Btrfly Net. Further, we validate this network for vertebrae labeling on in-house data supporting its use within a clinical diagnostic setting.

## 2. Materials and Methods

### Study Datasets

We worked with two datasets*: (a)* a public benchmarking dataset, $CSI_{label}$ and *(b)* a collection of two in-house datasets.

$CSI_{label}$ is a public dataset released in the Computational Spine Imaging (CSI) workshop during the 2014 Medical Image Computing and Computer Assisted Intervention (MICCAI) [5] by the Department of Radiology at the University of Washington. It consists of 302 spine-focused (i.e. tightly cropped) CT scans (officially split as 242 for training and 60 for testing) which include fractures, scans with and without contrast-enhancement, abnormal curvatures, and numerous (nearly 150 scans) post-operative cases with metal implants. The dataset has a mean voxel spacing of approximately $2 \times 0.35 \times 0.35$ $mm^3$ in the cranio-caudal × left-right × anterior-posterior directions and a mean dimension of (318±131) × (172±24) × (172±24) at 1-$mm^3$ isotropic resolution. Vertebrae centroids have been manually annotated as described by Glocker et al [5].

Our two in-house datasets consisted of 238 CT scans in total (178 for training and 60 for testing, with a 5-fold-cross validation). It is a collection of healthy and abnormal spines (e.g. osteoporosis, vertebral fractures, degeneration and scoliosis) with patients from 30 to 80 years old, collected for two previous retrospective studies [14,15]. Ethics approval was obtained from the local ethics committee of the faculty of medicine of the Technical University Munich for both studies. Due to the retrospective nature of these studies, the need for informed consent was waived. In brief, in Valentinitsch et al. [14], the authors established a normative atlas of the thoracolumbar spine in healthy, non-enhanced CT scans, collected between 2005 and 2014. In Valentinitsch et al. [15], the authors investigated texture analysis techniques for opportunistic osteoporosis diagnosis in contrast enhanced CT scans, collected from February 2007 to February 2008. For both studies, cases with metastasis and metal implants were excluded as they would change vertebral texture as well as biomechanical behavior and thus interfere with fracture prediction. Please note that these exclusion criteria are based on the older studies and not on the current one. In this study, we used n=65 non-contrast enhanced scans from [14] and n=173 contrast-enhanced scans from [15]. Overall, 20% of the dataset (46 scans) included parts of the ribcage, where 92 patients had fractured vertebrae. Annotations of the vertebrae were automatically derived from available segmentations, performed with an automated algorithm based on shape model matching [2]. They were verified by one radiologist (J. S. K) with more than 15 years of experience in spine imaging and corrected where necessary.

**CT Imaging**

All CT scans were performed either with a Philips 256-row or Siemens 128-row multidetector CT scanner and reconstructed using an edge-enhancing kernel. All scans were acquired with 120 kVp and an adaptive tube load. All subjects received i.v. contrast (Imeron 400; Bracco, Konstanz, Germany) with a delay of 70 s, a flow rate of 3 ml/s, and a body weight–dependent dose between 80 and 100ml. The mean voxel size of the dataset is approximately $0.7\times 2.75 \times 0.7$ mm$^3$ in the cranio-caudal $\times$ left-right $\times$ anterior-posterior directions with a mean size of $(565\pm111) \times (123\pm58) \times (401\pm140)$ mm$^3$.

**Spine Localization as a Pre-processing Stage.**

For efficiently handling data with diverse fields of views, localizing the spine was an important step that improved the 2D projections that all subsequent vertebrae labelling stages relied on. For localizing the spine, a 3D fully convolutional neural network was employed to regress Gaussian heat-maps centered at the vertebrae, followed by extracting a bounding box around the spine. This stage operated at very low resolution (4-mm isotropic), employing a light-weight, fully-convolutional U-network. The network architecture is described in supplementary Fig. E1 (Supplement). Fig. 1a shows examples of the predicted heatmaps and the extracted bounding boxes. With this spine localization-stage, we were able to extract a *localized MIP*, which is a MIP across only those slices that contain the spine. Localized MIP projections gave an unoccluded view of the spine (Fig. 1b).

**Vertebrae Labelling with the Btrfly Net**

Networks working in 3D are computationally intensive owing to their features vectors being of O($N^3$), where $N$ is one of the scan dimensions. As an alternative to working in a computationally demanding 3D domain, Sekuboyina et al. [11] proposed working with 2D sagittal and coronal MIPs by proposing a butterfly-shaped architecture termed *Btrfly Net*. As the Btrfly Net processes two-dimensional data, its feature dimensionality is limited to O($N^2$). This reduction in requirement of computational resources allows one to design deeper 2D networks with more convolutional layers leading to higher receptive fields. In this work, an improved version of the Btrfly Net is employed. Most importantly, we work with scans at 2-mm isotropic resolution, consequently increasing the receptive field and the representational capacity of the Btrfly Net (Fig. E2 [Supplement]).

**Adversarially Enforcing Spine's Shape onto the Btrfly Net**

Human anatomy usually follows certain structural rules, thus anatomic nomenclature has strong tacit assumptions (or priors), e.g. in case of the spine, vertebra L2 is almost always caudal to L1 and cranial to L3. Enabling a network to learn such priors results in anatomically consistent predictions. In our case, a secondary network (discriminator, **D**), which was trained along with the primary Btrfly Net, learned the spine's shape and forced the Btrfly Net's predictions to respect this shape. In Sekuboyina et al. [11], an energy-based autoencoder was used as a discriminator (EB-**D**). In this work, we improved EB-**D**'s architecture and compared it with a purely encoding architecture (W-**D**) [16]. The combination of the Btrfly Net and E-**D** was referred to as $Btrfly_{pe\text{-}eb}$ (denoting energy-based prior encoding) and that of the Btrfly Net and W-**D** was referred to as $Btrfly_{pe\text{-}w}$ (for Wasserstein-distance based prior encoding). The architectures, cost functions, and the adversarial training details of both the improved EB-**D** and proposed W-**D** are described in Supplemental Appendix E1. Note that EB-**D** is a fully-convolutional network and has a receptive field covering only a fixed part of the spine irrespective of the input scan dimension. On the other hand, W-**D** have a receptive field covering the full image owing to the dense connections in the architecture. We refer to these two as *local* and *global* receptive fields respectively and investigate the difference in their behavior.

**Inference**

Our algorithm included two components, spine localization and vertebrae labelling, as illustrated in Fig. 2. Given an multidetector CT scan, the localization stage predicts a heat-map (and a bounding box) indicating the location of the spine from which localized MIPs can be extracted and passed to the labelling stage. Each arm of the Btrfly Net labels the sagittal and coronal projections which are then fused by outer product to obtain the 3D vertebral locations. Note that the role of the discriminators ends with adversarially enforcing the spine prior onto the Brtfly Network during the training stage and are not a part of the inference path. The improvement in performance due to various components of our approach were assessed using a Wilcoxon signed-rank test assuming independence across scans.

### Performance Evaluation of Spine and Vertebral Detection

*Spine localization:* The performance of the localization stage is evaluated with two metrics defined as: *(a)* intersection-over-union between the actual and predicted bounding boxes and *(b)* detection rate, where a detection was successful if the corresponding intersection-over-union of greater than 50%. It was observed that this overlap suffices for our task of filtering out the obstructions.

*Vertebrae Labelling*: The labelling performance was evaluated using metrics defined in Glocker et al. [4] and Sekuboyina et al. [11] namely, identification rate (Id.rate), localization distance ($d_{mean}$ and $d_{std}$), precision ($P$), and recall ($R$). Briefly, a vertebra was correctly 'identified' if the precited vertebral location was the closest point to the ground truth location and was less than 20-mm away from it. The distance of this prediction was then recorded by the localization distance. $P$ and $R$ quantified the relevance of the predictions (eg. accurately labelling all vertebrae in the scan vs. labelling a vertebra that was not present in the scan).

### Statistical Analysis

Since the test sets are large with n=60 scans for the public and in-house datasets, we employed the non-parametric Wilcoxon signed-rank test for validating the statistical significance of the improvement in various performance measures due to our architectural modifications. These modifications included the fusion of the sagittal and coronal views in the Btrfly Net and the inclusion of prior-encoding components in $\text{Btrfly}_{pe\text{-}w}$ and $\text{Btrfly}_{pe\text{-}eb}$ Nets.

## 3. Results

### Successful Spine Localization within the $CSI_{label}$ and In-house Datasets

Localizing the spine in CT is a relatively simple task owing to the higher attenuation (measured by Hounsfield units) of the bone. The network only needs to isolate the spine from the rest of the skeletal structure. Table 1 records the performance of this stage on the two datasets in use. We obtain a mean intersection-over-union greater than 75% on both datasets. A detection rate of 100% on $CSI_{label}$ and 96% on the in-house dataset indicated a successful localization of the spine in all the cases. Note that a complete failure of localization might not be detrimental as it would only result in an occluded MIP. Labelling such a projection could be handled in the subsequent stages with appropriate augmentation if the failure rate is minimal.

### Btrfly Network Variants, $Btrfly_{pe\text{-}eb}$ and $Btrfly_{pe\text{-}w}$, Perform Refined Vertebrae Labelling compared to the Original Btrfly Network

We evaluated the vertebrae labelling performance in two stages*: (a)* on the public $CSI_{label}$ dataset, we evaluated the performance of the standalone labelling component and compared it with prior work (Table 2 and Table 3) *and (b)* on the in-house dataset, we deployed the combination of localization and labelling to evaluate the contribution of localizing the spine (Table 4). In order to be agnostic to initialization and since the train-test split is official, we reported the mean performance over three runs of training with independent initializations on the public dataset, while we performed a 5-fold cross validation on the in-house dataset to compensate for any dataset bias. All the improvements in the performance due to our contributions are statistically significant with $p < .05$ according to Wilcoxon signed-rank test.

*Contribution of Btrfly architecture:* In order to validate the importance of combining the views and processing them in the Btrfly Net, we compared its performance to a network setup working individually on the views without any such fusion of views. This setup was denoted by 'Cor.+Sag'. The architecture of each of these networks was similar to one arm of the Btrfly Net without the fusion of views. Due to this architectural modification, an increase of 1% in the Id.rate was observed between Cor.+Sag. Net and Btrfly Net as reported in Table 2.

*Effect of adversarial encoding:* Enforcing prior information into the Btrfly Net's training resulted in a 1-2% improvement in the Id.rate. Interestingly, we observed that $\text{Btrfly}_{\text{pe-eb}}$'s performance was marginally superior to $\text{Btrfly}_{\text{pe-w}}$'s (Table 2). This can be explained by the distinction between the *locally* encoding $\text{Btrfly}_{\text{pe-eb}}$ and a *globally* acting $\text{Btrfly}_{\text{pe-w}}$. This difference in encoding also explains the high variance of the Id.rate of $\text{Btrfly}_{\text{pe-w}}$ (standard deviation of 0.2 [$\text{Btrfly}_{\text{pe-eb}}$] vs. 1.2 [$\text{Btrfly}_{\text{pe-w}}$]). A qualitative comparison of the three variants in our experiments is shown in Fig. 3. The top row shows a use-case with successful labelling and the bottom rows shows an interesting failure use-case where the labelling fails due to the presence of an obstruction along the coronal view.

*Precision and Recall:* A Gaussian image predicted for any vertebra was counted as a final label if the response if higher than a threshold, $T$. Thus, $T$ controls the false-positive and true-positive label predictions of the network. Fig. 4 shows a precision-recall curve generated by varying $T$ between 0 to 0.8 in steps of 0.1, while Table 3 records the performance at the F1-optimal threshold. Despite not choosing a *recall-optimistic* threshold, our networks performed comparably well at an optimal-F1 threshold. Notice the over-arching nature of $\text{Btrfly}_{\text{pe-eb}}$ over others at all thresholds.

*Rigorous evaluation with respect to Id.rate:* As defined in Glocker et al. [4], a vertebra was accurately identified if it was the closest to the ground truth and less than 20-mm away. We denote this distance threshold by $d_{th}$. However, at $d_{th}$ = 20-mm could be a weak requirement, e.g. in case of a cervical vertebra which are very close to one another. Demonstrating the spatial precision of our localization, we performed a breakdown test with respect to the Id.rates by varying $d_{th}$ between 5-mm to 30-mm in steps of 5-mm. Fig. 5 shows the region-wise performance curves obtained for this variation across our setups. Notice the reasonably stable behavior of the curves until $d_{th}$ = 10-mm.

*Labelling with improved projections:* The performance of the $\text{Btrfly}_{\text{pe-eb}}$ architecture (chosen due to its superior performance) on naïve MIPs with that of localised MIPs on the in-house dataset is shown in Table 4. Observe an inferior Id.rate and high localization distances with naïve MIPs due the lack of visible vertebrae (Fig. 6). However, localized MIPs from the spine's bounding box results in about 4% gain in Id.rate.

## 4. Discussion

We proposed a generalizable pipeline to localize and identify vertebrae in multidetector CT scans that operates on appropriate 2D projections, unlike prior publications that work with 3D scans. Specifically, we incorporated an improved version of the Btrfly architecture in combination with a spine localization and a prior-learning stage.

It can be seen from the performance of the localization stage, with intersection-over-unions greater than 75%, that detecting the spine is an easier task. Note that the network localizes the spine better in the public dataset compared to the in-house dataset. This can be attributed to the composition of the datasets: $CSI_{label}$ is predominantly composed of thoracolumbar reformations with a uniform axial FOV centered around the spine, while the in-house dataset has a higher variety of scans in terms of axial and coronal FOVs. For the same reason, localizing the spine and extracting MIPs only across the slices containing the spine showed an improvement in labelling performance on the in-house dataset from 81% with naïve MIPs to 85% with localised MIPs. Such an improvement was not observed for scans from $CSI_{label}$ due to their relatively uniform FOVs. For the in-house dataset, the improvement can be attributed to the resulting homogeneity in the appearance of the MIPs owing to the localization irrespective of the scan content in the anteroposterior and lateral directions. Such data homogenization also resulted in a more stable learning.

In the module responsible for vertebral labelling, our algorithm contained two principle architectural components: *(a)* Fusion of the sagittal and coronal MIPs (using the Btrfly net) and *(b)* incorporation of anatomical prior information using the adversarial discriminators. This architecture, when trained end-to-end, resulted in Id.rate of 88% and localization distances of 6-mm on the public dataset, outperforming prior state-of-art [7,8,10]. The effectiveness of these architectural components can be analyzed in three stages: First, a setup working purely on 2D MIPs (Cor. + Sag.) readily outperformed the naïve 3D fully convolutional network proposed by Yang et al. [8]. This was mainly due to the depth of our setup, which can be afforded due to lesser computational load in 2D. Second, fusing the coronal and sagittal views increased the Id.rate by ~1%. This can be attributed to two reasons*: (a)* the advantage that one view gets key points in the other view, *and (b)* the combination of views causes the predictions of the Btrfly net to be spatially consistent between views. Third, the Btrfly net's prediction were made to respect an anatomical prior using adversarial discriminators (EB-**D** or W-**D**) as a proxy to the post-processing steps usually employed. For example, Yang et al. [8] used a learnt dictionary of vertebral centroids to correct a new prediction, Yang et al. [9] and Liao et al. [10] used a pre-trained recurrent neural network enforcing the vertebrae's sequence to the prediction. Substituting these secondary steps, our Btrfly net was trained adversarially. The second network in our case (the adversary) was neither pretrained nor was required during inference. The adversary provided a higher loss when the Btrfly network's prediction did not conform to a normal spine structure. This enforced the anatomical prior directly onto the Btrfly Net during training. It is interesting to note that on one of the cross-validation data splits of the in-house dataset, our method achieved a 100% vertebra identification in 34 of the 60 test scans. Moreover, of the remaining 26 scans, 7 scans had only one vertebrae mis-identified and 2 scans had only two vertebrae misidentified.

Delving deeper into prior-learning, EB-**D** whose receptive field covered a fixed part of a spine irrespective of the input FOV, enforced the spine's structure locally. In contrast, W-**D** enforced a global spine prior since the dense layers in it resulted in the receptive field covering the entire input scan. Note that adversarial learning in any form improved the Id.rate by 1-2%. However, $Btrfly_{pe-eb}$ marginally outperformed $Btrfly_{pe-w}$. This can be attributed to the fact that a local prior is easier to learn than a global one. Since spine multidetector CTs have highly varying FOVs, accurately learning a global prior was non-trivial for W-**D**. In EB-**D**'s case, the receptive field always covered about three vertebrae in lumbar region (and more elsewhere). Since these sub-regions are relatively similar across scans, it was relatively easier for EB-**D** to learn and enforce a local prior. This can also be observed in a lower standard deviation in the Id.rate of $Btrfly_{pe-eb}$ across initializations, indicating a more stable performance.

This study, however, has certain limitations stemming from design choices that need further investigation. First, concerning the proposed approach, the optimality of maximum intensity projections for labelling needs to be further studied so as to ascertain cases where they would break down. From the evaluations performed in this work, it seems to accurately label scoliotic spines, spines with metallic insertions, fractures etc. However, an organized analysis is lacking. Second, the independent training of the localization stage weakens our approach's 'end-to-end-trainable' feature. Moreover, the labelling performance depends on the accuracy of localization. When trained until convergence, localization seems to be accurate enough to aid the subsequent labelling stage. However, the robustness of the labelling stage to errors in spine localization is yet to be ascertained. Finally, the study presented here is retrospective and the computations have been performed on graphical processing units (GPU), where a forward pass takes less than 10 seconds. However, even if our model is lighter than its contemporaries, its demanding for a regular CPU and takes approximately 1 minute per scan. Hence, an effort towards further making the setup lighter is of interest. Finally, in consistence with the labels of $CSI_{label}$, we only considered 24 labels for C1-L5 and did not account for segmentation anomalies like L6 or transitional vertebra like a lumbarized S1.

In summary, this work presented a simple, yet efficient, algorithm for labelling vertebrae by arguing for processing the spine scans in two dimensions using an appropriate network architecture. Localizing the spine before labelling and forcing the labelling network to respect the spine's anatomical shape during prediction further improved the labelling performance. The entire setup is computationally lighter than its counterparts in literature, which is a step towards real-time clinical deployment.

## ACKNOWLEDGEMENTS

This work is supported by the European Research Council (ERC) under the European Union's 'Horizon 2020' research & innovation programme (GA637164–iBack–ERC–2014–STG). We also acknowledge NVIDIA Corporation's support with the donation of the Quadro P5000 used for this research.

## REFERENCES

1. S.Schmidt, J.Kappes, M.Bergtholdt, V.Pekar, S.Dries, D.Bystrov, and C. Schnorr. Spine detection and labeling using a parts-based graphical model. Information Processing in Medical Imaging. Springer, 2007.
2. T. Klinder, J. Ostermann, M. Ehm, A. Franz, R. Kneser, and C. Lorenz. Automated model-based vertebra detection, identification, and segmentation in ct images. Medical Image Analysis, vol. 13, no. 3, pp. 471–482, 2009.
3. J. Ma and L. Lu. Hierarchical segmentation and identification of thoracic vertebra using learning-based edge detection and coarse-to-fine deformable model. Comput. Vis. Image Underst., vol. 117, no. 9, pp. 1072–1083, 2013.
4. B. Glocker, J. Feulner, A. Criminisi, D. R. Haynor, and E. Konukoglu. Automatic localization and identification of vertebrae in arbitrary field- of-view ct scans. Medical Image Computing and Computer-Assisted Intervention. Springer, 2012.
5. B. Glocker, D. Zikic, E. Konukoglu, D. R. Haynor, and A. Criminisi. Vertebrae localization in pathological spine ct via dense classification from sparse annotations. Medical Image Computing and Computer- Assisted Intervention. Springer, 2013.
6. A. Suzani, A. Seitel, Y. Liu, S. Fels, R. N. Rohling, and P. Abolmaesumi. Fast automatic vertebrae detection and localization in pathological ct scans – a deep learning approach. Medical Image Computing and Computer-Assisted Intervention. Springer, 2015.
7. H. Chen, C. Shen, J. Qin, D. Ni, L. Shi, J. C. Y. Cheng, and P.-A. Heng. Automatic localin and identification of vertebrae in spine ct via a joint learning model with deep neural networks. Medical Image Computing and Computer-Assisted Intervention. Springer, 2015.
8. D. Yang, T. Xiong, D. Xu, Q. Huang, D. Liu, S. K. Zhou, Z. Xu, J. Park, M. Chen, T. D. Tran et al. Automatic vertebra labeling in large-scale 3d ct using deep image-to-image network with message passing and sparsity regularization. Information Processing in Medical Imaging. Springer, 2017.
9. D. Yang, T. Xiong, D. Xu, S. K. Zhou, Z. Xu, M. Chen, J. Park, S. Grbic, T. D. Tran, S. P. Chin et al.. Deep image-to-image recurrent network with shape basis learning for automatic vertebra labeling in large-scale 3d ct volumes. Medical Image Computing and Computer-Assisted Intervention. Springer, 2017.
10. H. Liao, A. Mesfin, and J. Luo. Joint vertebrae identification and localization in spinal ct images by combining short- and long-range contextual information. IEEE Transactions on Medical Imaging, vol. 37, no. 5, pp. 1266–1275, 2018.
11. A. Sekuboyina, M. Rempfler, J. Kukacka, G. Tetteh, A. Valentinitsch, J. S. Kirschke, and B. H. Menze. Btrfly net: Vertebrae labelling with energy-based adversarial learning of local spine prior. Medical Image Computing and Computer Assisted Intervention. Springer, 2018.

12. H. Ravishankar, R. Venkataramani, S. Thiruvenkadam, P. Sudhakar, and V. Vaidya. Learning and incorporating shape models for semantic segmentation. Medical Image Computing and Computer Assisted Intervention. Springer, 2017.

13. O. Oktay, E. Ferrante, K. Kamnitsas, M. P. Heinrich, W. Bai, J. Caballero, R. Guerrero, S. A. Cook, A. de Marvao, T. Dawes et al. Anatomically constrained neural networks (ACNN): application to cardiac image enhancement and segmentation. CoRR, vol. abs/1705.08302, 2017.

14. A. Valentinitsch, S. Trebeschi, E. Alarcón, T. Baum, J. Kaesmacher, C. Zimmer, C. Lorenz, J. S. Kirschke. Regional analysis of age-related local bone loss in the spine of a healthy population using 3D voxel-based modelling. Bone, Volume 103, 2017.

15. A. Valentinitsch, S. Trebeschi, J. Kaesmacher, , C. Lorenz, M. T. Loeffler, C. Zimmer, T. Baum, J. S. Kirschke. Opportunistic osteoporosis screening in multi-detector CT images via local classification of textures. Osteoporos Int. Jun 30 2019.

16. M. Arjovsky, S. Chintala, and L. Bottou. Wasserstein GAN. CoRR, vol. abs/1701.07875, 2017.

17. I. Gulrajani, F. Ahmed, M. Arjovsky, V. Dumoulin, and A. C. Courville. Improved training of wasserstein gans. Advances in Neural Information Processing Systems 30, 2017.

## Supplemental Appendix E1

In this section, we present the architectural and training details of various components in our approach in three parts: spine localization, Btrfly network, and prior-encoding using adversarial learning.

### *Spine Localization*

For localization, annotations were obtained from the available vertebral centroids by putting Gaussians at vertebral locations. The input was a 3D scan at low isotropic resolution of 4-mm. The output was also a 3D volume with a Gaussian at every vertebra location. We used a higher σ (=15) for wider Gaussians as annotations. The architecture was a light-weight U-Net as shown in Fig. S1. The loss used for training is a simple $\ell_2$ regression loss.

*Training:* For training, an Adam optimizer was employed with an initial learning rate of $5\times10^{-4}$. Convergence was tested on a held-out validation set consisting of 10% of the training data. Once trained, a bounding box is constructed out of the predicted heatmap using a fixed tolerance around the 'active' voxels in the prediction which indicate the presence of the spine.

### *Btrfly Net*

The detailed architecture of the Btrfly net is shown in Fig. S2. Similar to Sekuboyina et al. [11], the loss for the Btrfly net is a combination of cross-entropy and $\ell_2$ regression loss. Denoting $\boldsymbol{x}_{view}$ and $\boldsymbol{y}_{view}$, $view \in \{\text{sag}, \text{cor}\}$ to be the 2D projections of the image and the annotations respectively, the loss of the sagittal arm of the Btrfly Net was given by:

$$\mathcal{L}_{b,sag} = || \boldsymbol{y}_{sag} - \tilde{\boldsymbol{y}}_{sag} ||_2 + \omega H\left(\boldsymbol{y}^{\sigma}_{sag}, \tilde{\boldsymbol{y}}^{\sigma}_{sag}\right).$$

where $\tilde{\boldsymbol{y}}_{sag}$ is the prediction of the net's xz-arm, $H$ is the cross-entropy function, and $\boldsymbol{y}_{sag}^{\sigma} = \sigma(\boldsymbol{y}_{sag})$, the softmax excitation of the prediction. $\omega$ is the median frequency weight map over the occurrence of vertebral labels, incorporated for boosting the learning of less frequent vertebral classes. The loss for the yz-arm was similarly constructed resulting in the total loss of the Btrfly net:

$$\mathcal{L}_{btrfly} = \mathcal{L}_{b,sag} + \mathcal{L}_{b,cor}.$$

*Training:* Data for this phase consisted of MIPs across the two views. For training the network, an Adam optimiser was employed with an initial learning rate of $\lambda$ = 1×10$^{-3}$. $\lambda$ was decayed by a factor of 3/4$^{th}$ every 10k iterations till 0.2×10$^{-3}$. 10% of the training data was held out to ascertain training convergence.

***Prior learning using adversaries***

In this work, we also investigated the ability of two adversaries in aiding the labelling of vertebrae: an energy-based adversary and a Wasserstein-distance based adversary. Both the adversaries worked on the label maps ($\boldsymbol{y}_{view}$ or $\tilde{\boldsymbol{y}}_{view}$) and provided a discriminating signal specific to their architecture. This signal corresponded to 'real-ness' or 'fake-ness' of the input to the discriminator. Fig. E3 shows the arrangement of the discriminators with respect to the Btrfly Net and their detailed architecture.

***EB-D***: The discriminating signal here was the $\ell_2$ distance between the input $\boldsymbol{y}_{view}$ (or $\tilde{\boldsymbol{y}}_{view}$) and its reconstruction, $rec(\boldsymbol{y}_{view})$: $\mathcal{D}(\boldsymbol{y}_{view}) = E = ||\, \boldsymbol{y}_{view} - rec(\boldsymbol{y}_{view})||_2$. The total loss combined with the Btrfly net (generator, **G**) is as follows:

$$\mathcal{L}_D = \mathcal{D}(\boldsymbol{y}_{view}) + \max\big(0, m - \mathcal{D}(\tilde{\boldsymbol{y}}_{view})\big), \text{ and}$$

$$\mathcal{L}_G = \mathcal{D}(\widetilde{\boldsymbol{y}}_{view}) + \mathcal{L}_{b,view}$$

Thus, **D** learnt to better reconstruct 'real' label maps by minimising the reconstruction loss for a real sample. On the other hand, **G** learnt to predict better 'fake' label maps such that the discriminator gives a low reconstruction loss. $m$ in the equation above is a margin discouraging EB-**D** from learning to reconstruct fake or predicted samples. It was varied from 10 to 0 rapidly ($m$ was halved every 1000 iterations)

***W-D***: In case of WGAN, the **D** or **G** specific losses are not as intuitive as in the earlier case. However, the combined setup of the generator (Btrfly) and discriminator tried to minimise the Wasserstein distance between the distributions of real and generated label maps. We followed the loss design as in the improved Wasserstein GAN with gradient penalty [17] as below:

$$\mathcal{L}_D = \mathcal{D}(\widetilde{\boldsymbol{y}}_{view}) - \mathcal{D}(\boldsymbol{y}_{view}) + \lambda(||\nabla_{\widehat{\boldsymbol{y}}_{view}} \mathcal{D}(\boldsymbol{y}_{view})||_2 - 1)^2, \text{and}$$

$$\mathcal{L}_G = -\mathcal{D}(\widetilde{\boldsymbol{y}}_{view}) + \mathcal{L}_{b,view},$$

where $\widehat{\boldsymbol{y}}_{view} = \epsilon \boldsymbol{y}_{view} + (1-\epsilon)\widetilde{\boldsymbol{y}}_{view}$, for any $\epsilon \sim U[0,1]$. $\lambda$ (= 1.0 in our work) is a scalar weight for the regularisation term on the gradients.

## TABLES

Table 1: **Performance of the Spine Localization Stage**

| Measures | $CSI_{label}$ | In-house |
| --- | --- | --- |
| Mean IoU | 0.86 | 0.76 |
| Detection Rate (%) | 100 | 96.7 |

Note.— Intersection-over-union (IoU) (ratio between 0.0 to 1.0) records the intersection-over-union between the bounding boxes around the spine in the ground truth and the prediction. Detection rate captures the ratio of successful localizations in the dataset based on the intersection-over-union. Observe a high detection rate for both the public $CSI_{label}$ dataset and the in-house dataset. CSI= Computation Spine Imaging

Table 2: **Performance Comparison of our Approach with Prior Work**

| Approaches | Id.rate (in %) | | | | $d_{mean}$ ($d_{std}$) (in mm) | | | |
|---|---|---|---|---|---|---|---|---|
| | All | Cer. | Tho. | Lum. | All | Cer. | Tho. | Lum. |
| Chen et al [7] | 84.2 | 91.8 | 76.8 | 88.1 | 8.8 (13.0) | 5.1 (8.2) | 11.4(16.5) | 8.2 (8.6) |
| Yang et al [8] | 85 | 92 | 81 | 83 | 8.6 (7.8) | 5.6 (4.0) | 9.2 (7.9) | 11.0 (10.8) |
| Liao et al. [10] | 88.3 | 95.1 | 84.0 | 92.2 | 6.5 (8.6) | 4.5* (4.6) | 7.8 (10.2) | 5.6* (7.7) |
| Cor. + Sag. | 85.8±0.8 | 92.3±0.2* | 80.1±2.1 | 90.0 ±2.3 | 6.7 (5.4) | 5.8 (5.3) | 8.2 (7.4) | 7.2 (8.1) |
| Btrfly | 86.7±0.4 | 89.4±0.7 | 83.1±1.0 | 92.6±1.1 | 6.3 (4.0) | 6.1 (5.4) | 6.9 (5.5) | 5.7 (6.6) |
| $Btrfly_{pe-w}$ | 87.7±1.2 | 89.2±1.3 | 85.8±1.4 | 92.9±1.9* | 6.4 (4.2) | 5.8 (5.4) | 7.2 (5.7) | 5.6* (6.2) |
| $Btrfly_{pe-eb}$ | 88.5±0.2* | 89.9±0.2 | 86.2±0.4* | 91.4±1.7 | 6.2 (4.1)* | 5.9 (5.5) | 6.8 (5.9)* | 5.8 (6.6) |

Note.—The reported identification rate (Id.rate) is the mean across three runs of training with different initializations. However, the localization distances ($d_{mean}$ and $d_{std}$) are computed over the localization distances of all vertebrae in the three runs of training, i.e. over vertebrae in 3×60 test set scans. Improvement in Id.rate computed *per scan* across all variants is statistically significant in all runs ($p < .05$). Specifically, $Btrfly_{pe-eb}$'s performance gain is statistically significant with the lowest p-value (< .001). Values with an asterisk (*) indicate the best performer among the entries.

Table 3: **Precision and Recall Analysis**

| Approach | $T$ | $P$ (in %) | $R$ (in %) | F1 |
|---|---|---|---|---|
| Cor.+Sag. | 0.2 | 79.5 | 82.5 | 79.5 |
| Btrfly | 0.33 | 79.9 | 85.1 | 82.4 |
| $\text{Btrfly}_{\text{pe-w}}$ | 0.23 | 77.8 | 86.1 | 81.7 |
| $\text{Btrfly}_{\text{pe-eb}}$ | 0.23 | 80.2* | 87.9* | 83.4* |

Note.— Reported are the optimal mean $P$ and $R$ values based on F1 score, alongside mean of the F1-optimal threshold, $T$, for three runs. Details: $P$ = #hits/#predicted and $R =$ #hits/#actual, where #hits is the number of vertebrae satisfying the condition of identification as defined for Id.rate, #predicted is the vertebrae in the prediction, and #actual is the vertebrae actually present in the image. Values with asterisk (*) indicate the best performer among the entries.

Table 4: **Contribution of spine localization**

| Reformations | Id.rate (in %) | | | | $d_{mean}$ ($d_{std}$) (in mm) | | | |
|---|---|---|---|---|---|---|---|---|
| | All | Cer. | Tho. | Lum. | All | Cer. | Tho. | Lum. |
| Naïve MIP | 81.3±3.9 | 63.1±10.0 | 79.2±4.6 | 89.6±4.7 | 9.5 (8.2) | 14.1 (8.4) | 9.4 (7.6) | 9.5 (14.4) |
| Localised MIP | 85.1±1.2* | 65.7±6.7* | 83.7±3.1* | 92.0±2.0* | 8.1 (6.6)* | 13.4 (8.8)* | 8.1 (8.1)* | 7.7 (9.4)* |

Note.— The in-houses dataset contained a significant fraction of scans where the ribcage resulted in occluded maximum intensity projections. In such cases, localization of spine enabled extraction of 'localized maximum intensity projections' from only the spine region. Illustrated below is the advantage of offered by localized maximum intensity projections. Reported measures are obtained using the best performing architecture ($Btrfly_{pe\text{-}eb}$) from Table 2, averaged over five folds of cross-validation. Values with an asterisk (*) indicate the best performer among the entries. Id.rate = identification rate

## FIGURE LEGENDS

Figure 1: Spine localization and its necessity illustrated on a data sample from the in-house dataset: (a) Ground truth and predicted localization maps along with extracted bounding boxes (blue - actual, green - predicted), plotted on sagittal and coronal slices of a scan. (b) A naïve maximum intensity projection (MIP) of a scan containing the ribcage completely occludes the spine (left). This can be handled by extracting a localized MIP of the same scan using the bounding box of the localized spine (right).

Figure 2: An overview of our labelling approach. Also illustrated is the spine localization stage which is, in some cases, necessary for our approach to be generalizable to any clinical CT scan including complete vertebrae.

Figure 3: Qualitative comparison of our network architectures showing two cases form the public dataset with one successful (top, sagittal view) and one failed (bottom, coronal view) labelling. Observe, Cor.+Sag. and Btrfly nets label mostly in presence of spatial information. However, the prior-encoded $\text{Btrfly}_{\text{pe-eb}}$ and $\text{Btrfly}_{\text{pe-w}}$ hallucinate prospective vertebral labels despite no image information (labels on shaded gray on coronal view). Additionally, $\text{Btrfly}_{\text{pe-eb}}$ tries to retain the order of vertebral labels.

Figure 4: Precision-recall curve with F1 isolines, illustrating the effect of the threshold $T$ during inference. For any $T$, $\text{Btrfly}_{\text{pe-eb}}$ offers a better trade-off between precision, $P$, and recall, $R$.

Figure 5. Region wise variation of Id.rates (y-axis) for different values of the distance threshold ($d_{th}$, x-axis) considered as a positive identification.

Figure 6. A Localized maximum intensity projections (MIPs) alongside a naive maximum intensity projections from the same scan from the in-house dataset using the $Btrfly_{pe\text{-}eb}$ network. Observe occlusions in naive MIP and their lack thereof in localized MIP, resulting in an improved labelling performance.

***Supplementary figures:***

Figure E1: **Localization network architecture**: Kernel sizes: All convolution kernels except the last layer are 3×3×3. Last convolutional kernel is 1×1×1. Transposed convolution kernels are 4×4×4. Average pooling kernels are 2×2×2 with a stride of 2.

Figure E2: **Btrfly-Net architecture**: The xz- (blue) and the yz-arms (yellow) correspond to the sagittal and coronal views. In upscaling path, kernel size -×- denotes to two different kernel sizes: 4×4 transposed convolution followed by 3×3 convolution.

Figure E3: (a) The arrangement of the discriminators with respect to the Btrfly net. (b) (b) The architecture of the Wasserstein-distance-based discriminator (W-**D**). (c) The architecture of the energy-based discriminator (EB-**D**), giving an $\ell_2$ reconstruction error as output. In both the architectures, Leaky ReLU and batch normalization is employed after every layer except the last.

# FIGURES

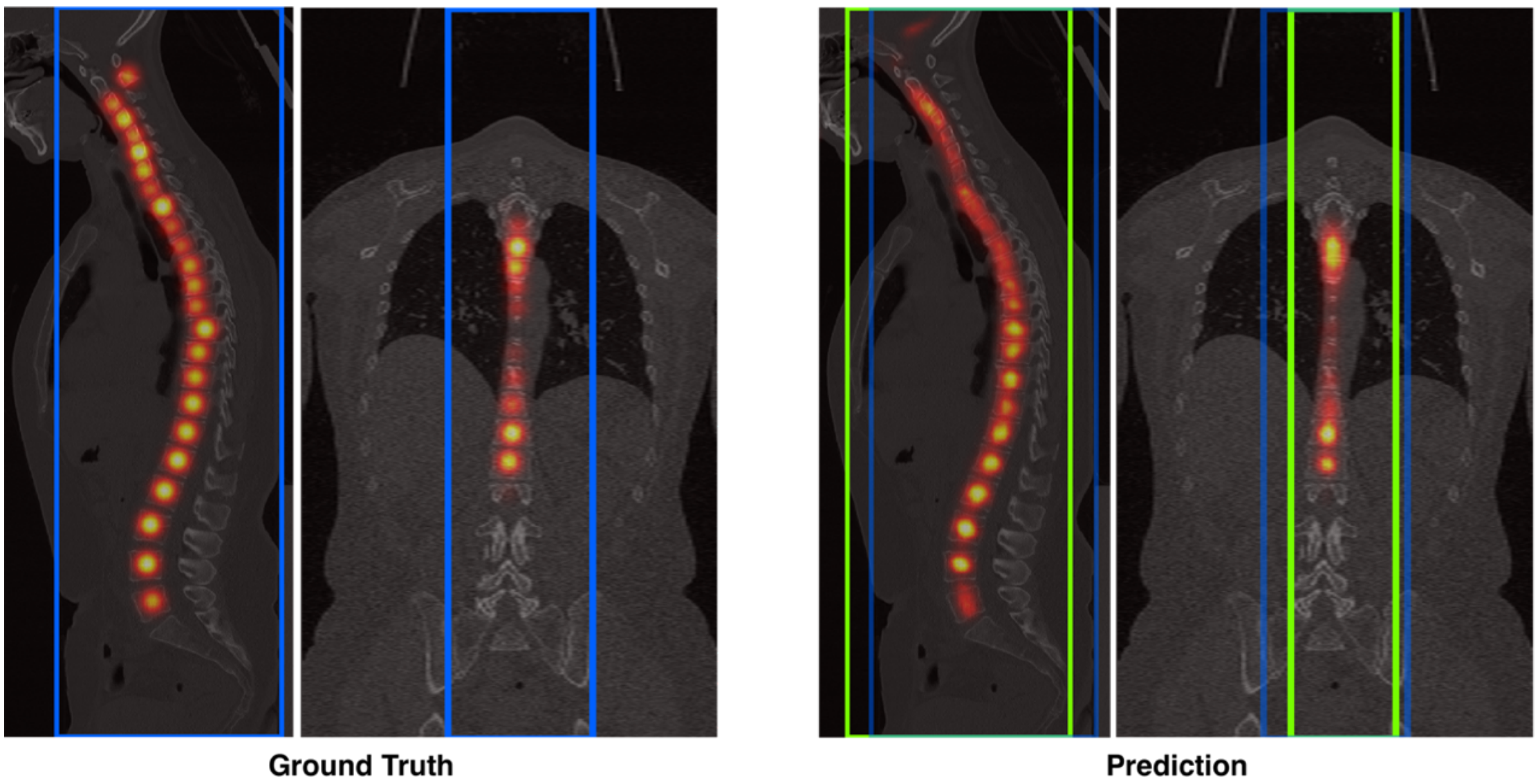

(a)

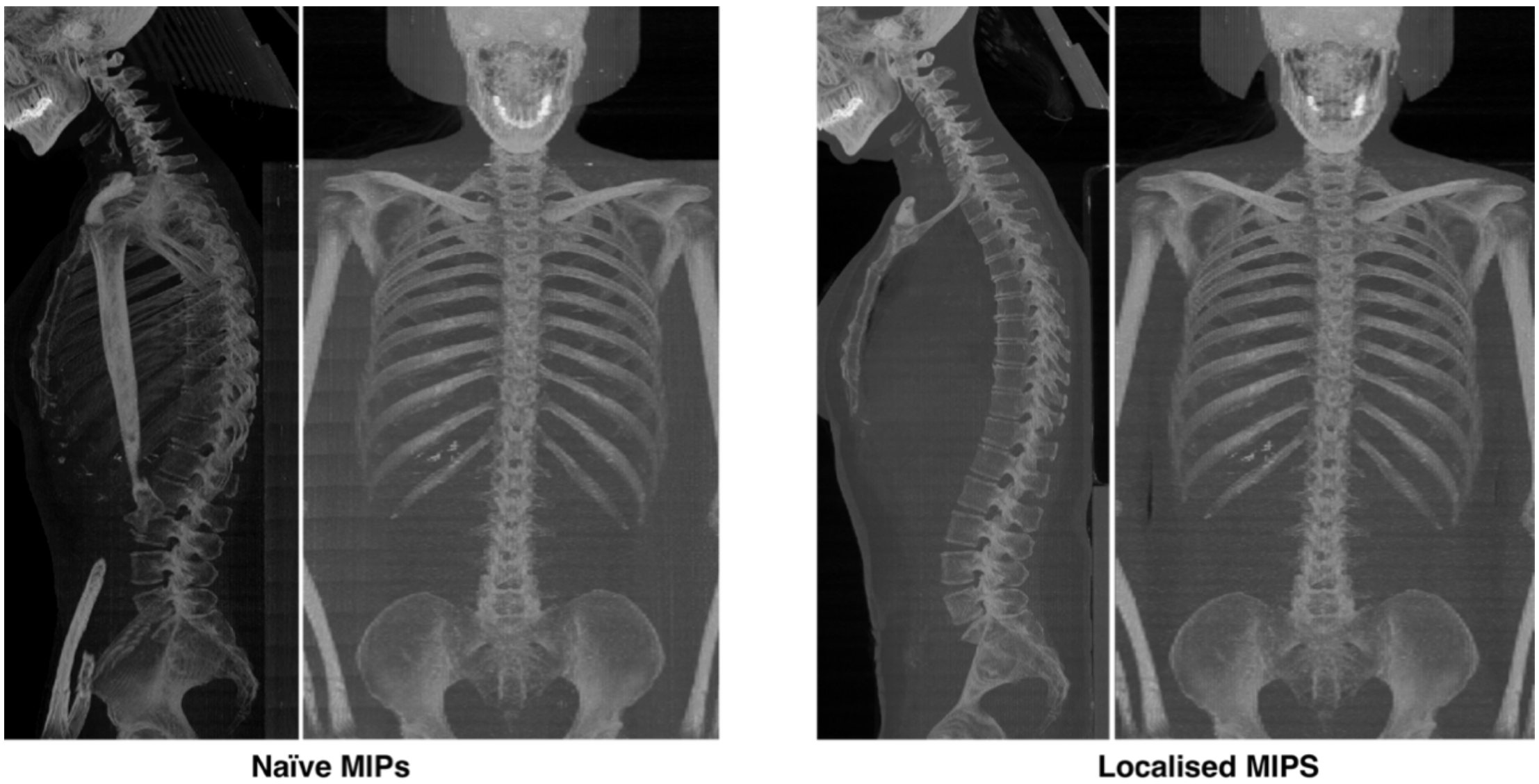

(b)

Figure 1: Spine localization and its necessity illustrated on a data sample from the in-house dataset: (a) Ground truth and predicted localization maps along with extracted bounding boxes (blue - actual, green - predicted), plotted on sagittal and coronal slices of a scan. (b) A naïve maximum intensity projection (MIP) of a scan containing the ribcage completely occludes the spine (left). This can be handled by extracting a localized MIP of the same scan using the bounding box of the localized spine (right).

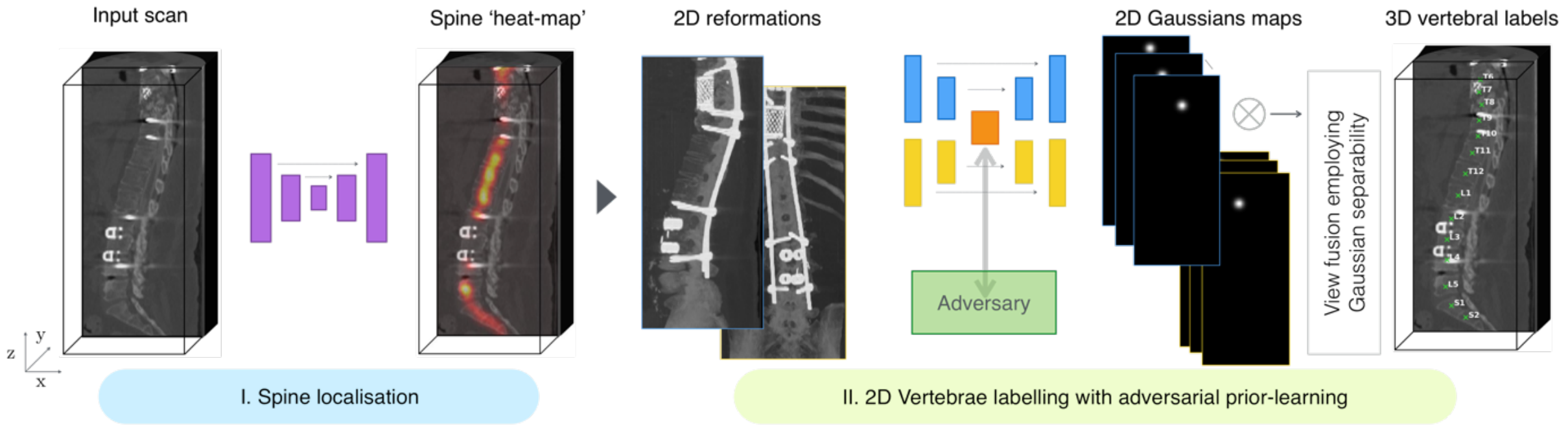

Figure 2: An overview of our labelling approach. Also illustrated is the spine localization stage which is, in some cases, necessary for our approach to be generalizable to any clinical CT scan including complete vertebrae.

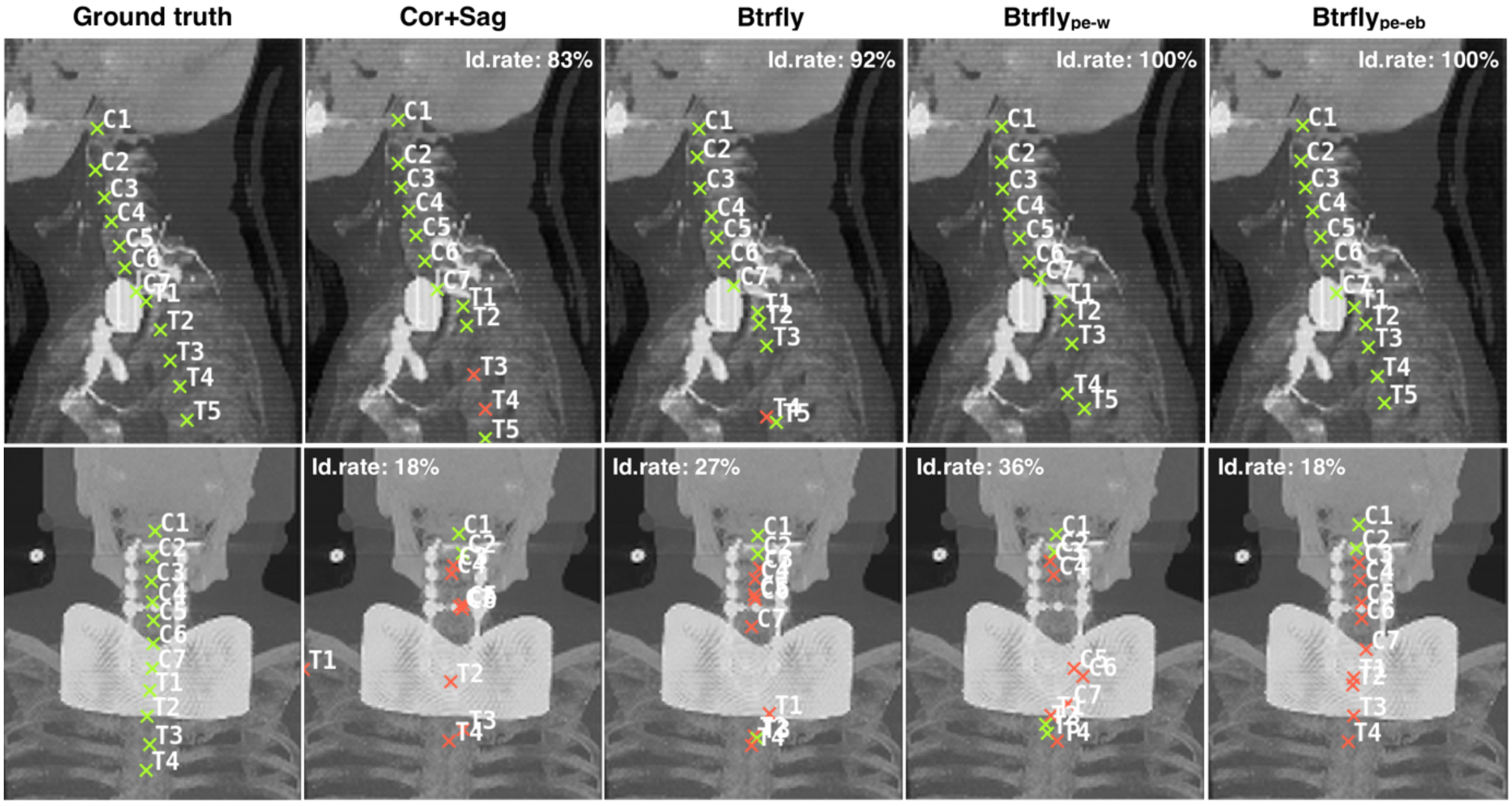

Figure 3: Qualitative comparison of our network architectures showing two cases form the public dataset with one successful (top, sagittal view) and one failed (bottom, coronal view) labelling. Observe, Cor.+Sag. and Btrfly nets label mostly in presence of spatial information. However, the prior-encoded $Btrfly_{pe-eb}$ and $Btrfly_{pe-w}$ hallucinate prospective vertebral labels despite no image information (labels on shaded gray on coronal view). Additionally, $Btrfly_{pe-eb}$ tries to retain the order of vertebral labels.

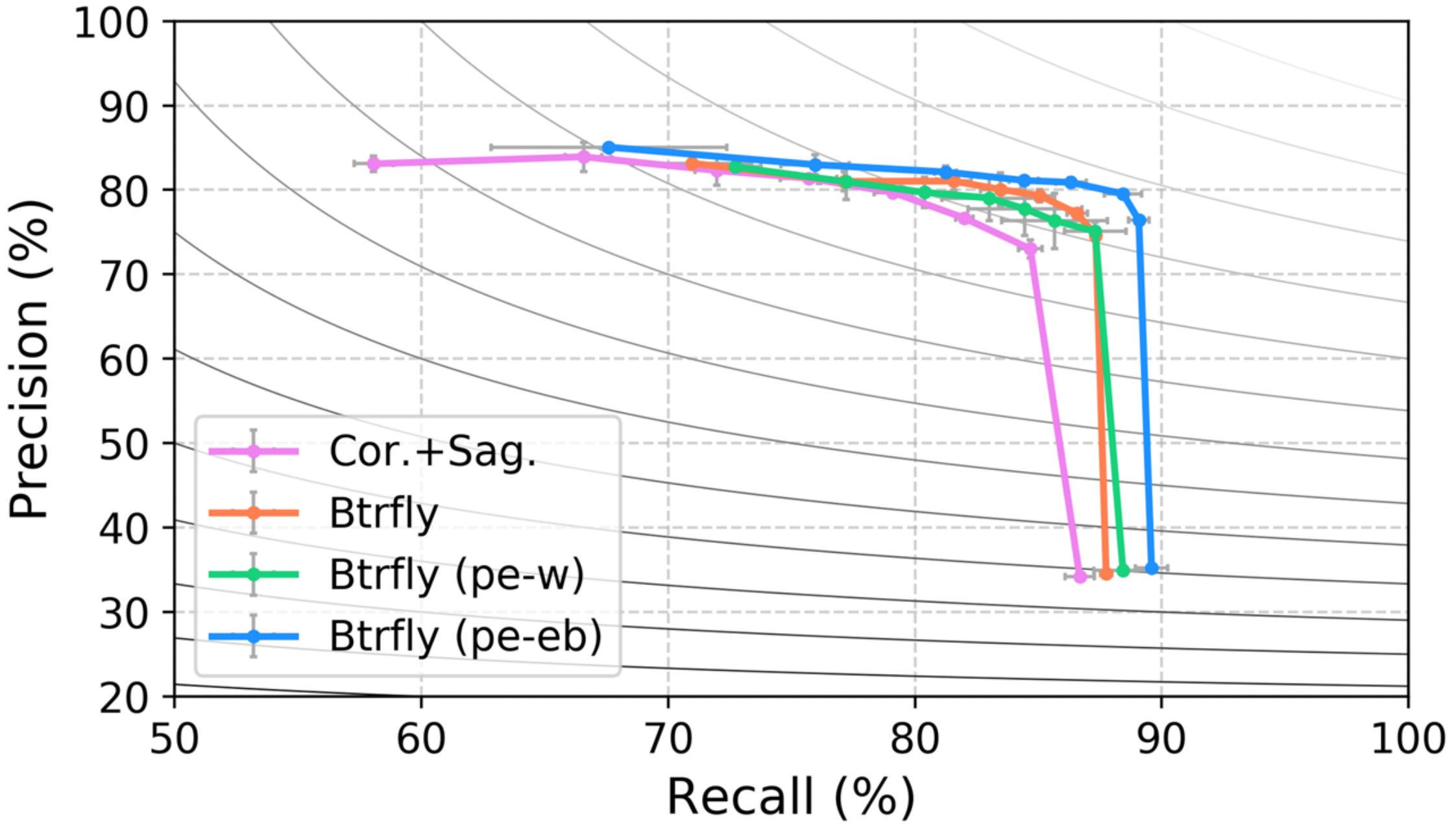

Figure 4: Precision-recall curve with F1 isolines, illustrating the effect of the threshold $T$ during inference. For any $T$, Btrfly$_{\text{pe-eb}}$ offers a better trade-off between Precision $P$ and Recall $R$.

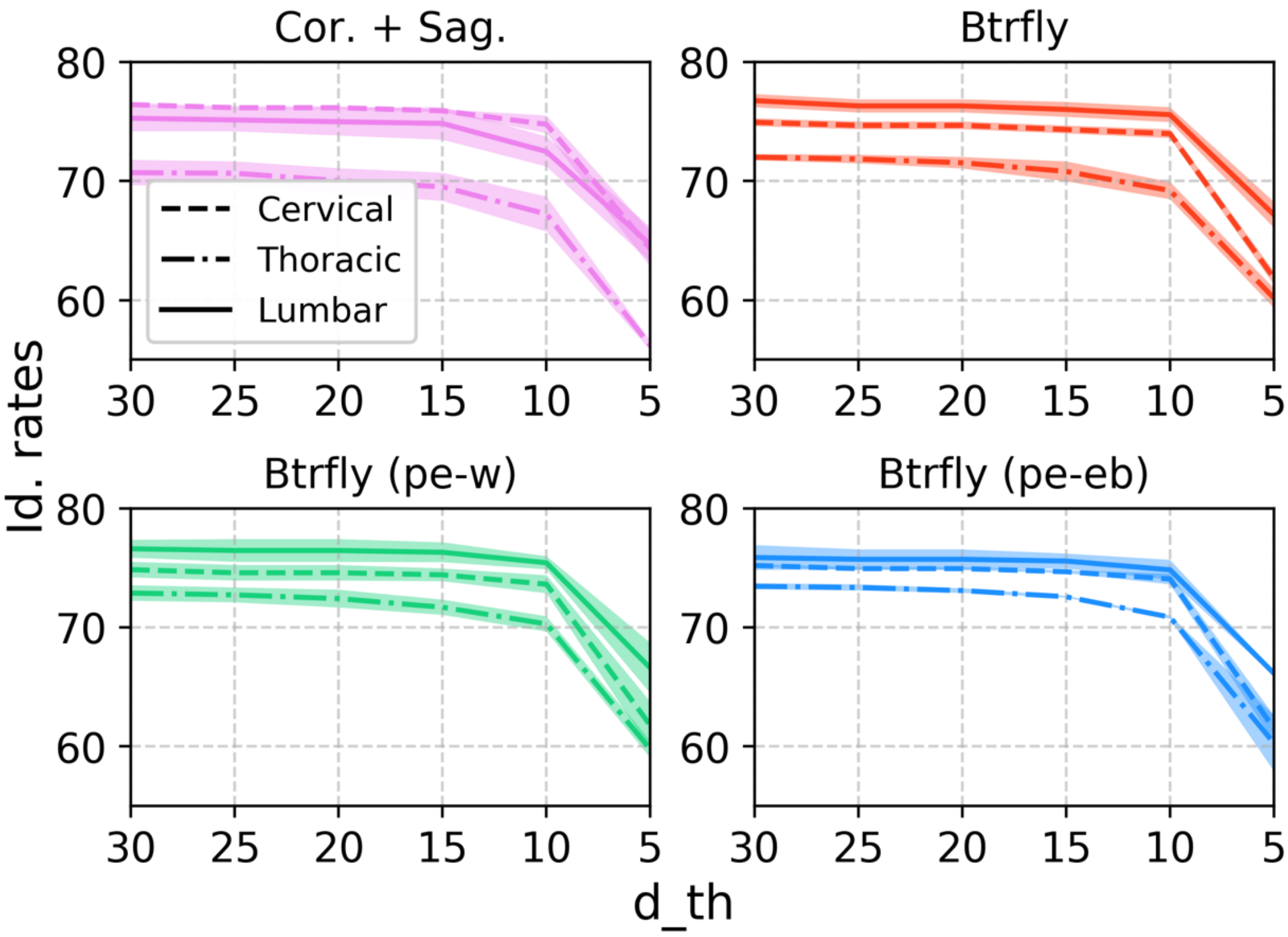

Figure 5. Region wise variation of Id.rates (y-axis) for different values of the distance threshold ($d_{th}$, x-axis) considered as a positive identification.

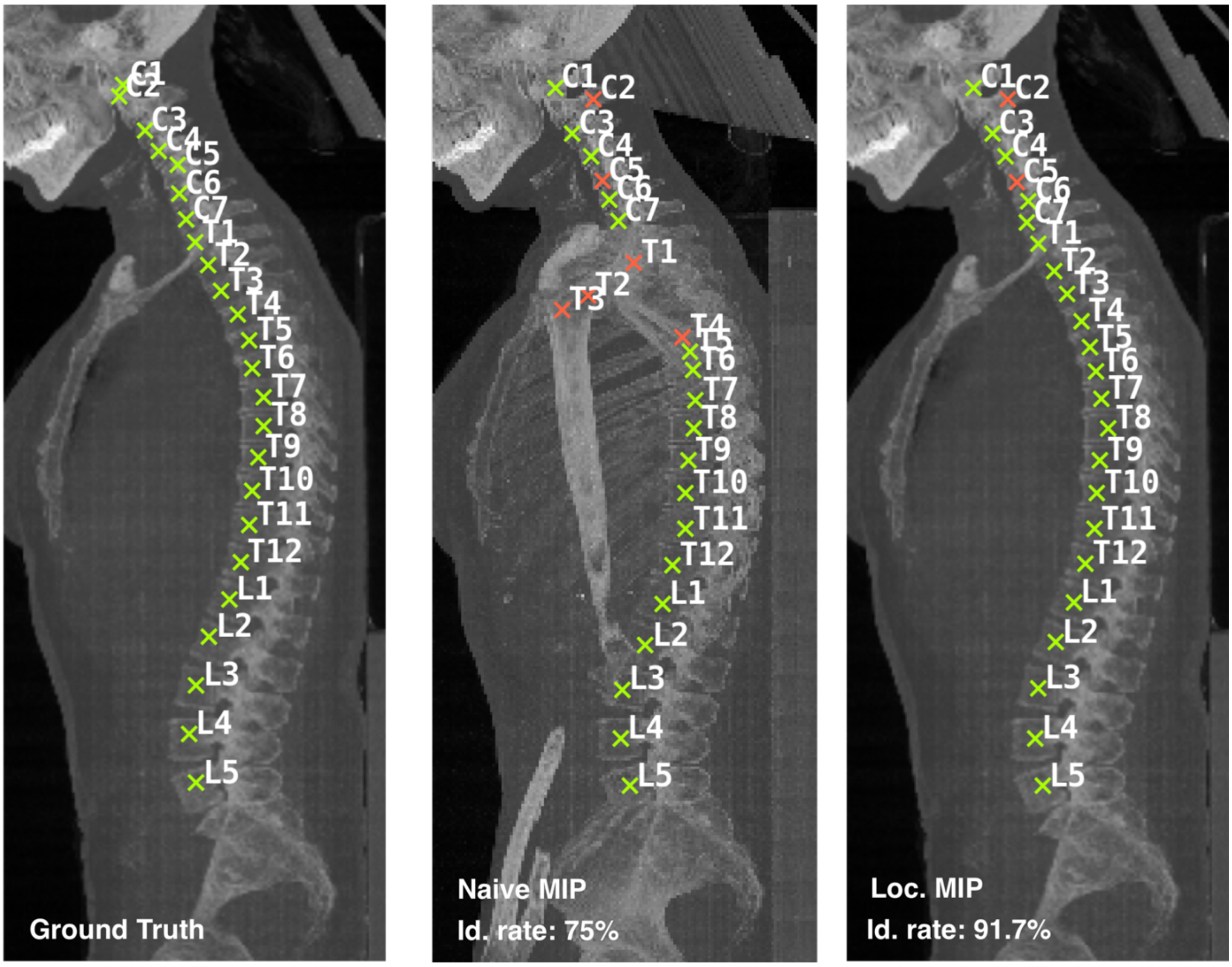

Figure 6. A Localized maximum intensity projections (MIPs) alongside a naive maximum intensity projections from the same scan from the in-house dataset using the $Btrfly_{pe\text{-}eb}$ network. Observe occlusions in naive MIP and their lack thereof in localized MIP, resulting in an improved labelling performance.

# SUPPLEMENTARY FIGURES

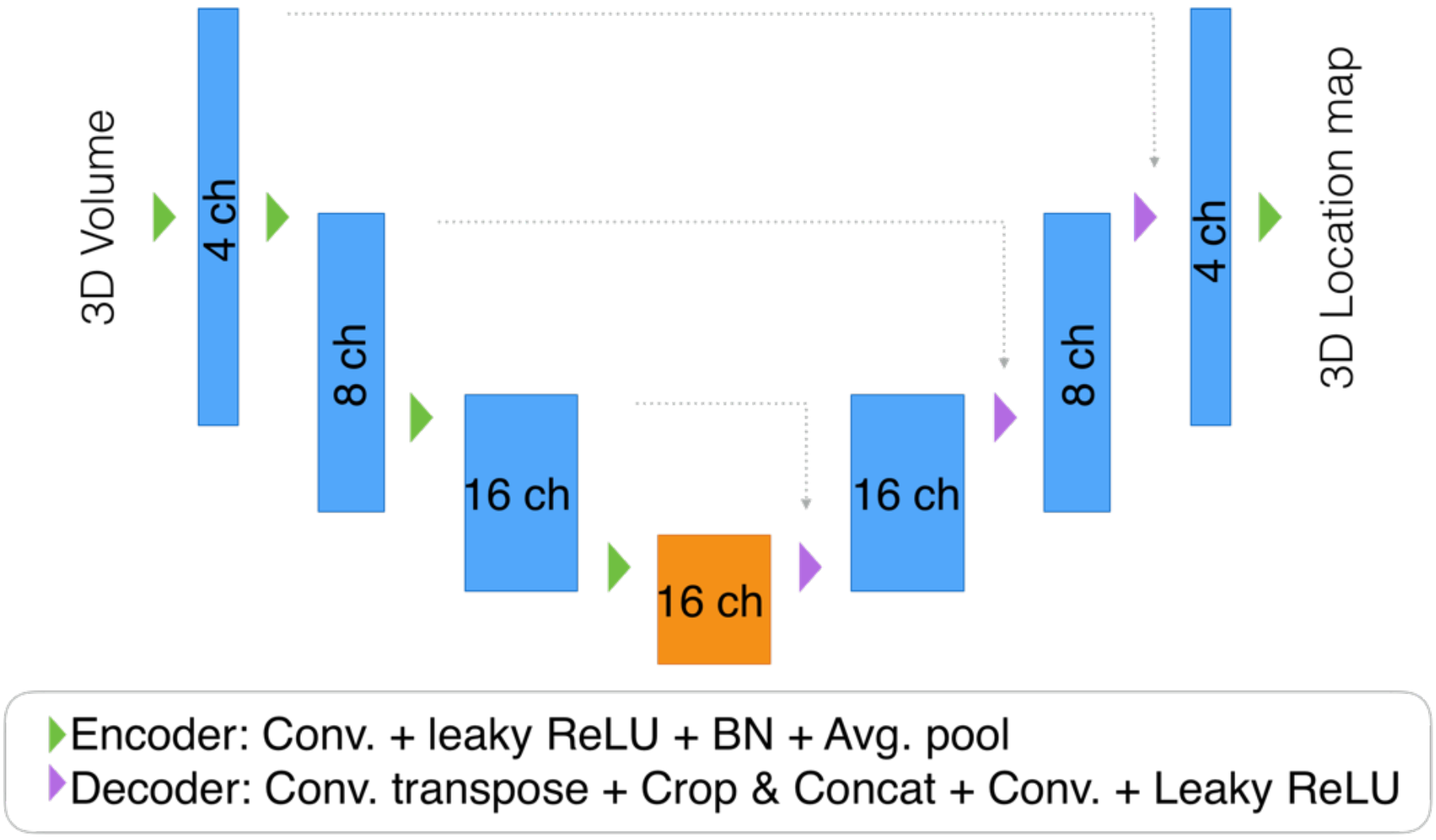

Figure E1: **Localization network architecture**: Kernel sizes: All convolution kernels except the last layer are 3×3×3. Last convolutional kernel is 1×1×1. Transposed convolution kernels are 4×4×4. Average pooling kernels are 2×2×2 with a stride of 2.

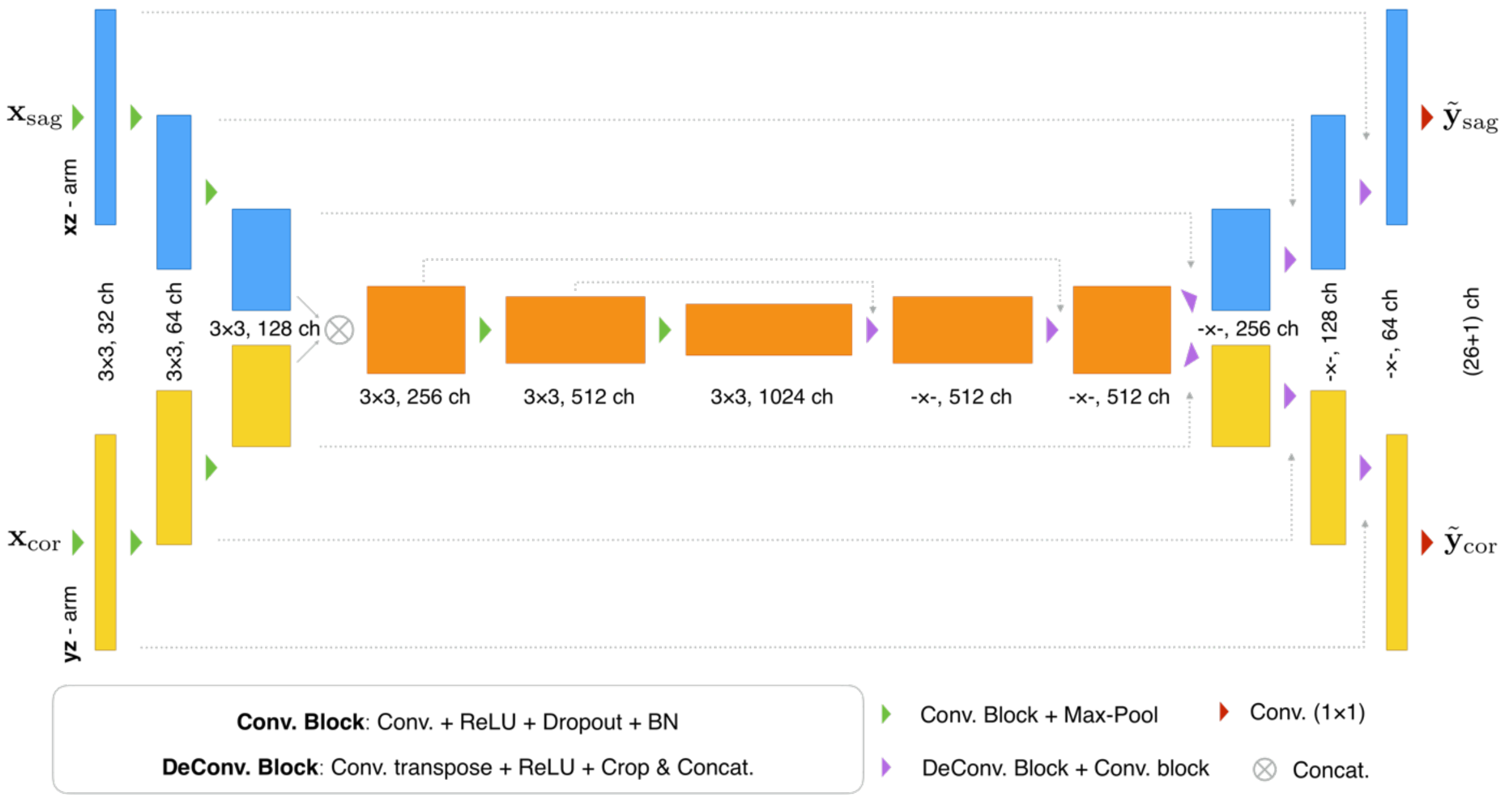

Figure E2: **Btrfly-Net architecture**: The xz- (blue) and the yz-arms (yellow) correspond to the sagittal and coronal views. In upscaling path, kernel size -×- denotes to two different kernel sizes: 4×4 transposed convolution followed by 3×3 convolution.

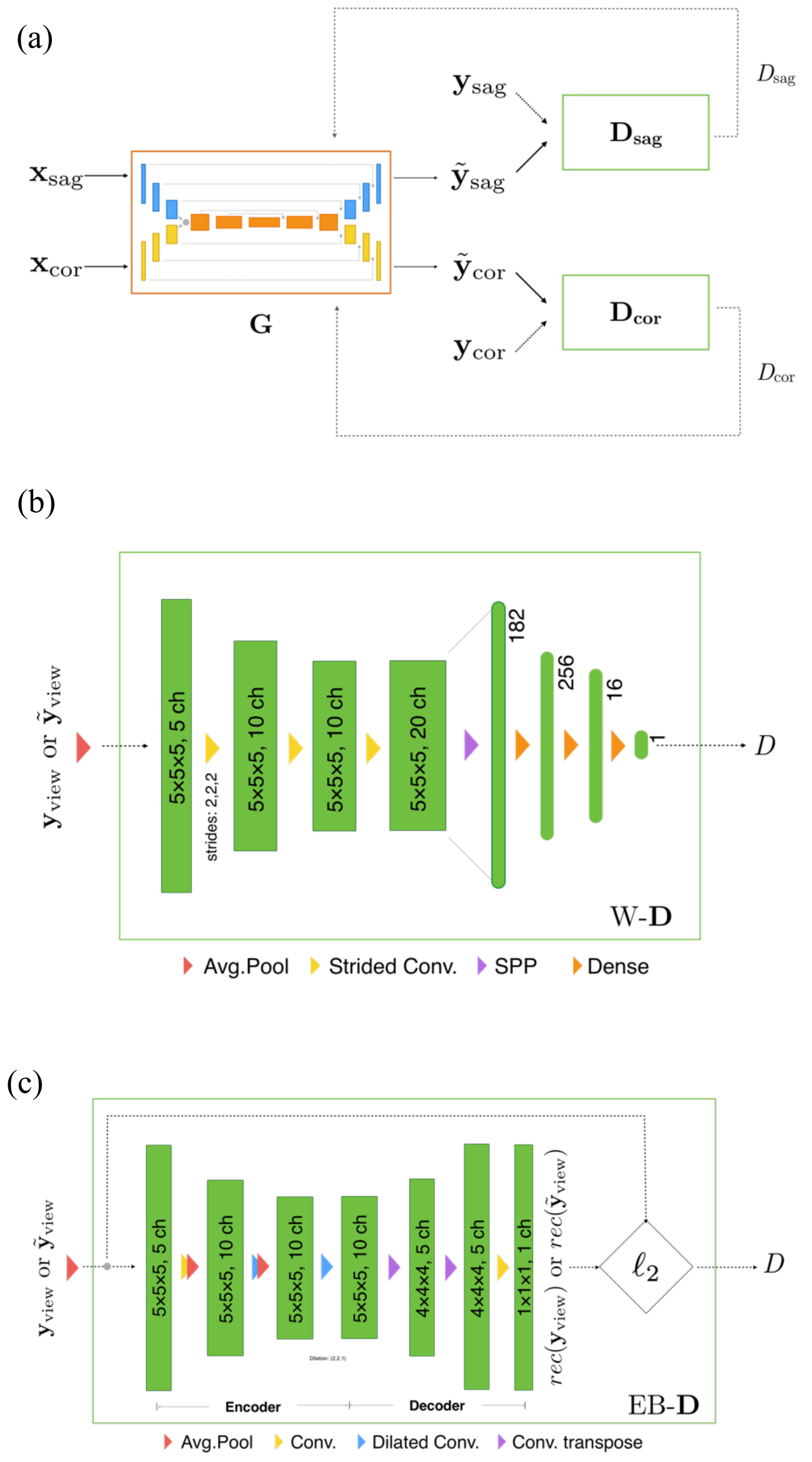

Figure E3: (a) The arrangement of the discriminators with respect to the Btrfly net. (b) (b) The architecture of the Wasserstein-distance-based discriminator (W-**D**). (c) The architecture of the energy-based discriminator (EB-**D**), giving an $\ell_2$ reconstruction error as output. In both the architectures, Leaky ReLU and batch normalisation is employed after every layer except the last.